\let\oldyear\year 
\documentclass{ieeeaccess} 
\let\year\oldyear 

\usepackage[utf8]{inputenc}
\usepackage[T1]{fontenc}
\usepackage{textcomp}
\usepackage[autostyle]{csquotes}
\usepackage[english]{babel}

\usepackage{amsmath, amssymb, bm}
\usepackage{mathtools}  

\makeatletter
\AtBeginDocument{
  \DeclareMathVersion{bold}
  \SetSymbolFont{operators}{bold}{T1}{times}{b}{n}
  \SetSymbolFont{NewLetters}{bold}{T1}{times}{b}{it}
  \SetMathAlphabet{\mathrm}{bold}{T1}{times}{b}{n}
  \SetMathAlphabet{\mathit}{bold}{T1}{times}{b}{it}
  \SetMathAlphabet{\mathbf}{bold}{T1}{times}{b}{n}
  \SetMathAlphabet{\mathtt}{bold}{OT1}{pcr}{b}{n}
  \SetSymbolFont{symbols}{bold}{OMS}{cmsy}{b}{n}
  \renewcommand\boldmath{\@nomath\boldmath\mathversion{bold}}
}
\makeatother

\usepackage{graphicx}
\usepackage[export]{adjustbox}  
\usepackage{booktabs}          
\usepackage{array}             
\usepackage{float}             

\usepackage{enumitem}
\setlist[itemize]{leftmargin=*,labelsep=0.4em,topsep=0pt,partopsep=0pt}
\setlist[enumerate]{leftmargin=*,labelsep=0.4em,topsep=0pt,partopsep=0pt}

\usepackage{cite}
\usepackage[colorlinks=true, allcolors=blue]{hyperref}  

\usepackage{siunitx}
\usepackage[acronym]{glossaries}  
\makeglossaries                

\newacronym{f1}{F\textsubscript{1}}{harmonic mean of precision and recall}
\newacronym{ffn}{FFN}{feed-forward network}
\newacronym{cnn}{CNN}{convolutional neural network}
\newacronym{auc}{AUC}{area under the ROC curve}
\newacronym{roc}{ROC}{receiver-operating characteristic}
\newacronym{pr}{PR}{precision-recall}
\newacronym{mlp}{MLP}{multilayer perceptron}
\newacronym{bce}{BCE}{binary cross-entropy}
\newacronym{pmffnn}{PMFFNN}{parallel multi-path feed-forward neural network}
\newacronym{lstm}{LSTM}{long short-term memory network}
\newacronym{cnnone}{1D-CNN}{one-dimensional convolutional neural network}
\newacronym{gpu}{GPU}{graphics processing unit}
\newacronym{cpu}{CPU}{central processing unit}
\newacronym{vpd}{VPD}{vapour-pressure deficit}
\newacronym{modis}{MODIS}{Moderate-Resolution Imaging Spectroradiometer}
\newacronym{viirs}{VIIRS}{Visible Infrared Imaging Radiometer Suite}
\newacronym{ndvi}{NDVI}{normalized difference vegetation index}
\newacronym{gpp}{GPP}{gross primary productivity}
\newacronym{spi}{SPI}{standardized precipitation index}
\newacronym{ml}{ML}{machine learning}
\newacronym{dl}{DL}{deep learning}

\usepackage{algorithmic}
\usepackage{algorithm}  

\usepackage{xspace}     
\usepackage{microtype}  
\usepackage{etoolbox}   
\def\BibTeX{{\rm B\kern-.05em{\sc i\kern-.025em b}\kern-.08em
    T\kern-.1667em\lower.7ex\hbox{E}\kern-.125emX}}

\begin{document}

\history{Date of publication April 15, 2025, date of current version April 15, 2025. This research was financed entirely with the authors' own funds.}
\doi{10.1109/ACCESS.2025.XXXXXXX}

\title{Deep Learning with Pretrained 'Internal World' Layers: A Gemma 3-Based Modular Architecture for Wildfire Prediction}
\author{\uppercase{Ayoub Jadouli}\authorrefmark{1} and \uppercase{Chaker El Amrani}\authorrefmark{1}, \IEEEmembership{Member, IEEE}}
\address[1]{Computer Science and Smart Systems, Faculty of Sciences and Technology, Abdelmalek Essaâdi University, 90000 Tangier, Morocco (e-mail: ajadouli@uae.ac.ma)}

\markboth
{Jadouli \headeretal: Preparation of Papers for IEEE ACCESS}
{Jadouli \headeretal: Preparation of Papers for IEEE ACCESS}

\corresp{Corresponding author: Ayoub Jadouli (e-mail: ajadouli@uae.ac.ma).}
\begin{abstract}
Deep learning models, especially large Transformers, carry substantial "memory" in their intermediate layers---an \emph{internal world} that encodes a wealth of relational and contextual knowledge. This work harnesses that internal world for wildfire occurrence prediction by introducing a modular architecture built upon Gemma 3, a state-of-the-art multimodal model. Rather than relying on Gemma 3's original embedding and positional encoding stacks, we develop a custom feed-forward module that transforms tabular wildfire features into the hidden dimension required by Gemma 3's mid-layer Transformer blocks. We freeze these Gemma 3 sub-layers---thus preserving their pretrained representation power---while training only the smaller input and output networks. This approach minimizes the number of trainable parameters and reduces the risk of overfitting on limited wildfire data, yet retains the benefits of Gemma 3's broad knowledge. Evaluations on a Moroccan wildfire dataset demonstrate improved predictive accuracy and robustness compared to standard feed-forward and convolutional baselines. Ablation studies confirm that the frozen Transformer layers consistently contribute to better representations, underscoring the feasibility of reusing large-model mid-layers as a learned internal world. Our findings suggest that strategic modular reuse of pretrained Transformers can enable more data-efficient and interpretable solutions for critical environmental applications such as wildfire risk management.
\end{abstract}


\titlepgskip=-21pt

\maketitle
\section{Introduction}
Wildfires are among the most devastating natural disasters, causing severe damage to ecosystems, economies, and human lives worldwide \cite{abatzoglou2016impact,chuvieco2010development}. As climate variability increases and human influences grow, accurately forecasting wildfire occurrences has become an urgent global challenge. Traditional fire-risk rating systems, while useful, often fail to capture the intricate interactions between meteorological, ecological, and anthropogenic factors \cite{parisien2009environmental,cortez2007data}.
Recent advances in deep neural networks (DNNs) have opened new avenues for modeling these complex interdependencies. In particular, large language models (LLMs) and large vision models (LVMs) have emerged as powerful tools that internalize vast amounts of knowledge---spanning human logic, mathematical principles, and scientific theory---through pretraining on diverse datasets \cite{brown2020language,wei2022emergent}. These models develop what can be considered an internal world: a rich repository of latent knowledge and reasoning capabilities embedded within their middle layers. This phenomenon raises an intriguing question: Could we leverage these pretrained "internal worlds" for specialized environmental prediction tasks like wildfire forecasting?
Our research has progressively built towards answering this question through a series of studies addressing wildfire prediction in Morocco. Jadouli and El Amrani \cite{jadouli2024advanced} first developed an advanced dataset specifically curated from multisource observations in Morocco, addressing local geographical and climatic characteristics. Building on this foundation, they proposed a Parallel Multi-path Feed Forward Neural Network (PMFFNN) \cite{jadouli2024pmffnn} that effectively reduces complexity in long columnar datasets, offering an efficient approach to processing heterogeneous environmental data.

This initial work revealed a critical insight: the temporal memory of past environmental states is crucial for accurate wildfire prediction. Subsequent research by Jadouli and El Amrani incorporated physical principles into the deep learning pipeline through a physics-embedded model \cite{jadouli2025physics} that integrates statistical mechanics into neural networks, thereby enhancing the interpretability of wildfire risk assessments. Further refinements demonstrated that integrating multisource spatio-temporal data via ensemble models and transfer learning significantly improves forecast accuracy \cite{jadouli2024enhancing}, and that bridging physical entropy theory with deep learning provides a hybrid approach that leverages satellite data for more interpretable environmental modeling \cite{jadouli2023bridging}.

This progressive research path has culminated in a key insight: pretrained models with their rich internal representations could potentially serve as sophisticated "memory modules" for wildfire prediction. While traditional approaches train models from scratch on domain-specific data, they miss the opportunity to leverage the knowledge and reasoning capabilities already embedded in large pretrained models. Our work addresses this gap by introducing a novel architecture that injects the middle layers of the state-of-the-art Gemma 3 model as an "internal world" module within a wildfire prediction network.
By carefully designing input projection layers to map tabular wildfire features into Gemma's hidden representation space, and bypassing its original embedding and positional encoding components, we access the rich knowledge captured in Gemma's pretrained transformer layers. These frozen layers---which encapsulate extensive knowledge from diverse domains---function as a sophisticated feature processor that enhances wildfire prediction without requiring extensive retraining. This approach not only reduces the number of trainable parameters and mitigates overfitting but also potentially improves model interpretability by grounding predictions in the established knowledge embedded within Gemma's weights.
The primary contributions of this paper are threefold:
Architectural Innovation: We propose a novel hybrid architecture that incorporates the middle layers of a large pretrained model (Gemma 3) as a modular "internal world" within a wildfire prediction network, demonstrating a new paradigm for leveraging pretrained knowledge in environmental modeling.
Memory-Enhanced Prediction: By exploiting the rich representational capacity of pretrained transformer layers, our model captures complex temporal dependencies and environmental interactions critical for wildfire prediction, addressing the limitations of traditional approaches that lack sophisticated memory mechanisms.
Efficient Transfer Learning: We demonstrate that by treating pretrained model layers as fixed knowledge modules, we can achieve superior performance with fewer trainable parameters, providing an efficient alternative to full fine-tuning or training specialized models from scratch.
The remainder of this paper is organized as follows. Section II reviews related work in wildfire prediction and the application of transfer learning in environmental modeling. Section III details our methodology, including our approach to integrating Gemma's "internal world" into the wildfire prediction architecture. Section IV presents experimental results and comparative analyses, and Section V concludes with implications and future research directions.

\section{Related Work}
This section reviews three key areas relevant to our proposed approach: (1) wildfire prediction methodologies, (2) representational power of pretrained model middle layers, and (3) modular reuse for transfer learning.

\subsection{Wildfire Prediction Methodologies}
Wildfire prediction methods have evolved from traditional weather index systems to advanced data-driven models. Early operational models like the Canadian Fire Weather Index \cite{van1977fire} and the National Fire Danger Rating System by the U.S. Forest Service \cite{deeming1972national} effectively utilized meteorological data but were limited by regional specificity and insufficient ecological interaction modeling \cite{parisien2009environmental}. Recent machine learning approaches integrate diverse datasets to achieve superior performance. Cortez and Morais \cite{cortez2007data} first utilized data mining techniques on meteorological variables, while comprehensive reviews like Jain \textit{et al.} \cite{jain2020review} summarized various machine learning applications. Deep learning, especially Convolutional Neural Networks (CNNs), has significantly advanced wildfire prediction by leveraging spatial information from satellite imagery \cite{zhang2019deep}. Notably, FireCast \cite{radke2019firecast}, a CNN-based multimodal system, showcased the predictive power of deep learning.

More recently, transformer-based models demonstrated exceptional capabilities by effectively capturing temporal dependencies. Lahrichi \textit{et al.} \cite{lahrichi2024predicting} introduced a Transformer-based Swin-Unet model that substantially outperformed conventional methods by utilizing multi-day historical inputs. Zhu \textit{et al.} \cite{zhu2023shapley} reinforced this trend, illustrating that transformer architectures yield higher accuracy and interpretability, underscoring the importance of temporal context in wildfire modeling.

\subsection{Author Trajectory and Workflow Integration}
Our previous research provides a structured progression toward the proposed hybrid architecture:

\paragraph{Scene-Level Ignition Proxies (2022).} We demonstrated the efficacy of CNN-based scene classification for identifying potential ignition sources in remote-sensing imagery \cite{jadouli2022human}, highlighting the role of contextual cues beyond meteorology.

\paragraph{Physics-Anchored Complexity (2023).} To enhance interpretability, we incorporated Boltzmann--Gibbs entropy into a 3-D CNN, developing a physics-informed complexity layer translating imagery into meaningful complexity metrics \cite{jadouli2023bridging}.

\paragraph{Data Foundation and Benchmarking (2024).} Addressing data scarcity, we created the first comprehensive multisource wildfire dataset for Morocco, benchmarking various machine learning approaches including CNNs and LSTMs \cite{jadouli2024advanced}.

\paragraph{Efficient Tabular Processing (2024).} The introduction of the Parallel Multi-path Feed Forward Neural Network (PMFFNN) facilitated efficient handling of wide tabular datasets, significantly reducing computational overhead \cite{jadouli2024pmffnn}.

\paragraph{Transfer and Ensemble Gains (2024).} Pre-training on global fire datasets and employing ensemble methods significantly improved predictive accuracy, demonstrating the advantages of leveraging external knowledge \cite{jadouli2024enhancing}.

\paragraph{Physics-Embedded Tabular Models (2025).} Embedding entropy layers within PMFFNN yielded a balanced model that effectively combined interpretability and nonlinear representation capabilities \cite{jadouli2025physics}.

\paragraph{Hybrid Spatial--Temporal Synthesis (2025).} A hybrid architecture comprising FFN, 1D-CNN, and LSTM achieved superior validation accuracy, demonstrating the effectiveness of combining spatial filters with temporal memory \cite{jadouli2025hybrid}. Incorporating sinusoidal positional encodings within FFN reduced the performance gap relative to full transformer models \cite{jadouli2025ffnpe}.

\subsection{Transformer Foundations and Gemma-3 Architecture}
Transformers, introduced by Vaswani \textit{et al.} \cite{vaswani2017attention}, replaced recurrence with multi-head attention, significantly improving long-range dependency modeling. Pretrained transformers like BERT \cite{devlin2019bert} demonstrated universal representation capabilities transferable across tasks. TabTransformer \cite{huang2021tabtransformer} further illustrated transformers' flexibility for structured, non-textual data.

Gemma-3 \cite{gemma3tech2025}, Google's lightweight multimodal transformer derived from Gemini 2.0, incorporates grouped-query attention, gated feed-forward layers, and a memory-efficient architecture handling extensive context windows (128K tokens). Gemma-3's design choices---rich latent knowledge from multimodal pretraining, extensive temporal memory, and compact deployment---make it ideally suited as an "internal world" module within our hybrid wildfire prediction architecture.

\subsection{Representation Power of Middle Layers}
Research has consistently shown the representational depth captured by intermediate layers in pretrained models. For instance, BERT's intermediate layers progressively encode syntactic and semantic information, highlighting their critical role in contextual understanding \cite{rogers2020primer}. Wallat \textit{et al.} \cite{wallat2021bertnesia} identified that significant knowledge resides within BERT's intermediate layers, suggesting a layered knowledge structure. Similarly, Tenney \textit{et al.} \cite{tenney2019bert} demonstrated BERT's layers implicitly form a processing pipeline from basic syntactic to advanced semantic relations.

In vision transformers, intermediate layers capture mid-level abstractions essential for higher-level tasks, with transformers demonstrating distinct hierarchical representations compared to CNNs \cite{raghu2021vision}. Such findings support the notion of pretrained models possessing "internal world models," enabling sophisticated reasoning through intermediate representations \cite{anthropic2022mechanistic}.

\subsection{Modular Reuse and Transfer Learning}
Traditional transfer learning approaches typically involve fine-tuning entire pretrained models or utilizing their final layers as fixed feature extractors \cite{pan2010survey}. Modular reuse, treating neural network components as reusable modules, is gaining attention due to its potential efficiency and flexibility.

Notably, He \textit{et al.} \cite{he2019rethinking} demonstrated CNN backbones provide substantial benefits for cross-domain tasks. Similarly, Lahrichi \textit{et al.} \cite{lahrichi2024predicting} showed pretrained vision transformers greatly enhanced wildfire prediction tasks. Imtiaz \textit{et al.} \cite{imtiaz2022neural} introduced modular neural architectures, highlighting minimal performance degradation from module reuse. Andreas \textit{et al.} \cite{andreas2016neural} pioneered dynamically composable neural modules for flexible task adaptation.

Recently, adapters, small trainable modules inserted into pretrained networks, proved highly effective for fine-tuning without substantial performance loss \cite{houlsby2019parameter}. Our approach adopts the inverse paradigm: inserting a pretrained internal world module into task-specific architectures, thus preserving general pretrained knowledge while achieving adaptability.

Collectively, these foundational works inform our novel architecture leveraging Gemma-3's pretrained internal layers, proposing a hybrid solution optimized for robust wildfire prediction.

\section{Methodology Overview}
\label{sec:methodology}
The goal of this work is to predict the daily occurrence of wildfires by
\emph{re-using} the middle transformer layers of \textbf{Gemma 3-1B}
as a \emph{frozen internal world} inside a lightweight, task--specific
network.  This section (roughly the first one-third of the full
Methodology) sets up the problem, describes the dataset, \textbf{fully
specifies the architecture---including every tensor shape and
hyper--parameter------and justifies each design choice}.  Training strategy,
baselines, and evaluation appear later in the sections.

\subsection{Problem Formulation}
\label{sec:formulation}
Let a sliding window of width \(w=30\)\ d stack \(d=276\)\ tabular
features (meteorology, vegetation indices, etc.).  The resulting sample
tensor is
\(
\mathbf{X}_t\in\mathbb{R}^{w\times d}.
\)
The task is binary:
\(
y_t\!=\!1
\)
if at least one MODIS/VIIRS fire pixel is detected in the same grid
cell on day \(t{+}1\); otherwise \(y_t=0\).
A model
\(f_\theta\)
with parameters \(\theta\) returns a probability
\(
\hat y_t=f_\theta\!\bigl(\mathbf{X}_t\bigr)\in[0,1].
\)
Training minimises the class-weighted cross-entropy
\(
\mathcal{L}_{\text{BCE}}
\)
with positive weight \(w_1{=}2.0\).

\subsection{Dataset Schema and Pre-processing}
\label{sec:data}
\begin{table}[!t]
\centering
\caption{Morocco wildfire dataset (2010--2022).}
\label{tab:data_schema}
\begin{footnotesize}
\begin{tabular}{@{}ll@{}}
\toprule
\textbf{Column}  & \textbf{Description} \\ \midrule
acq\_date   & Timestamp (YYYY--MM--DD)                              \\
is\_fire    & Target label: 1=fire, 0=nofire                 \\
vegetation  & NDVI, EVI, GPP, 15-day lags                         \\
moisture    & Soil-moisture layers, VCI, SPI, lags                \\
temp        & $T_\mathrm{max}$, $T_\mathrm{min}$, LST, VPD, lags  \\
precip      & Daily / cumulative rainfall metrics, lags           \\
wind        & Wind speed, direction, Fire-Weather Index sub-indices\\
topo        & Elevation, slope, aspect(static)                   \\ \bottomrule
\end{tabular}
\end{footnotesize}
\end{table}

\vspace{2pt}
\noindent
\textbf{Temporal split.}
Samples before \texttt{2022-01-01} are used for training,
2022 forms the validation year---replicating operational deployment.

\textbf{Balancing.}
Each split is undersampled so that
\(|\text{pos}|{:}|\text{neg}|=1{:}1\).

\textbf{Standardisation.}
Every numeric column is z-scored with \(\mu,\sigma\) from training
data only.

\textbf{Window assembly.}
For each day \(t\) we collect
\(
[\mathbf{x}_{t-w+1};\!\dots;\mathbf{x}_t]
\)
giving
\(
\mathbf{X}_t\in\mathbb{R}^{30\times276}.
\)

\subsection{Architectural Design Principles}
\label{sec:design}
The network is decomposed into three \emph{mutable--immutable} blocks:

\begin{enumerate}
\item \textbf{Input adaptation (trainable)}  
      --- four-branch \emph{Parallel Multi-path FFN} (PMFFNN) that
      projects tabular data into Gemma's hidden space
      (\(H=1152\)).
\item \textbf{Internal world (frozen)}  
      ---layers8--9 of Gemma3-1B's decoder; 14.7M parameters held
      constant.
\item \textbf{Output adaptation (trainable)}  
      ---two-layer MLP that maps the frozen representation to a scalar
      logit.
\end{enumerate}
\subsection{Complete TensorFlow and Hyperparameters}
\label{sec:arch_flow}
\begin{figure}[!t]
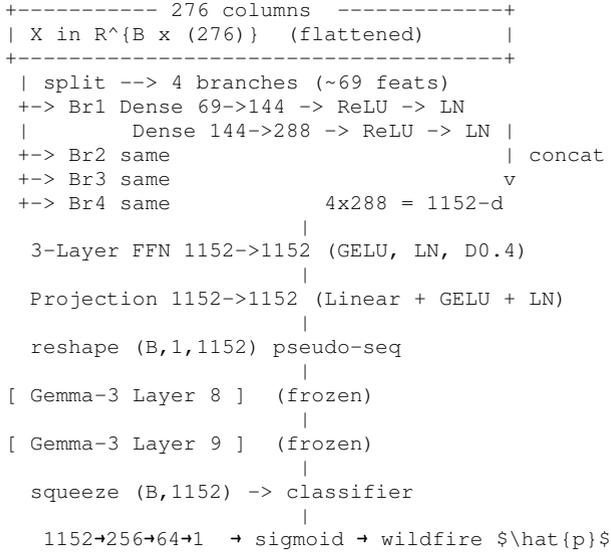
                     
  \centering
  {\footnotesize\ttfamily              
\begin{verbatim}
+----------- 276 columns  -------------+
| X in R^{B x (276)}  (flattened)      |
+--------------------------------------+
 | split --> 4 branches (~69 feats)
 +-> Br1 Dense 69->144 -> ReLU -> LN
 |        Dense 144->288 -> ReLU -> LN |
 +-> Br2 same                          | concat
 +-> Br3 same                          v
 +-> Br4 same            4x288 = 1152-d
                       |
  3-Layer FFN 1152->1152 (GELU, LN, D0.4)
                       |
  Projection 1152->1152 (Linear + GELU + LN)
                       |
  reshape (B,1,1152) pseudo-seq
                       |
[ Gemma-3 Layer 8 ]  (frozen)
                       |
[ Gemma-3 Layer 9 ]  (frozen)
                       |
  squeeze (B,1152) -> classifier
                       |
   1152→256→64→1  → sigmoid → wildfire $\hat{p}$
\end{verbatim}}
  \caption{Text-based schematic of the \emph{Internal-World Model}.
  Brackets mark \textbf{frozen} Gemma layers; solid boxes are trainable.}
  \label{fig:iwm_schema}
\end{figure}

Figure~\ref{fig:iwm_schema} shows the end-to-end ASCII schematic; all
symbols are defined in Table~\ref{tab:symbols}. Parameter counts are
summarized in Table~\ref{tab:param_breakdown}.

\begin{table}[H]
\centering
\caption{Symbol definition for architectural hyperparameters.}
\label{tab:symbols}
\begin{tabular}{@{}ll@{}}
\toprule
Symbol & Meaning / Value \\
\midrule
\(B\)          & Mini-batch size (128) \\
\(w\)          & Temporal window length (30 d) \\
\(d\)          & Raw feature count (276) \\
\(H\)          & Gemma hidden size (1152) \\
\(h_b\)        & Branch output size (\(H/4=288\)) \\
\(n_{\text{heads}}\) & Attention heads in Gemma slice (8) \\
\(L_{\text{IW}}\)    & \# frozen transformer blocks (2) \\
Drop-in        & 0.4 (projection FFN), 0.3 (classifier) \\
LR\(_{\text{proj}}\) & 1 × 10\(^{-3}\) \\
LR\(_{\text{cls}}\)  & 5 × 10\(^{-4}\) \\
\bottomrule
\end{tabular}
\end{table}
\begin{table}[!t]
\centering
\caption{Trainable parameters per block (Gemma slice frozen).}
\label{tab:param_breakdown}
\footnotesize      
\begin{tabular}{@{}lcr@{}}
\toprule
\textbf{Block} & \textbf{Trainable?} & \textbf{Params (M)} \\
\midrule
Four branches (PMFFNN)   & Yes & 3.4\footnotemark[1] \\
Cross-branch FFN (3-layer) & Yes & 4.0\footnotemark[2] \\
Projection head          & Yes & 1.3 \\
\textbf{Internal world}  & No  & 14.7 (frozen) \\
Classifier (2-layer)     & Yes & 0.34\footnotemark[3] \\
\midrule
\textbf{Total}          & --  & 21.7 M (trainable $\approx$ 5.0 M) \\
\bottomrule
\end{tabular}
\footnotetext[1]{\(4\,[69\!\times\!144+144\!\times\!288]\).}
\footnotetext[2]{\(3H^2,\,H{=}1152\).}
\footnotetext[3]{\(1152\!\times\!256+256\!\times\!64+64\!\times\!1\).}
\end{table}

\subsection{Rationale for Key Hyper-Parameters}
\label{sec:rationale}
\begin{itemize}
\item \textbf{Hidden size (\(H=1152\)).}  
      Matches Gemma-3-1B; avoids projection overhead inside the frozen
      block and preserves the pre-learned distribution of activations.
\item \textbf{Branch width (\(h_b=288\)).}  
      Ensures four-way concatenation exactly reaches \(H\); wider
      branches gave marginal gains at double the parameter cost.
\item \textbf{Attention heads (\(n_{\text{heads}}=8\)).}  
      Native to Gemma-3-1B; we preserve head count to maintain
      pre-training priors about key--query dimensionality.
\item \textbf{Dropout 0.4.}  
      A grid search in \(\{0.2,0.3,0.4,0.5\}\) selected 0.4 as the best
      trade-off between regularisation and underfitting for 5 M
      trainable weights.
\item \textbf{Learning-rate split.}  
      Empirically, the projection network requires faster adaptation,
      thus it receives a learning rate 2× that of the shallower
      classifier.
\end{itemize}

\subsection{Optimisation Protocol and Computational Considerations}
\label{sec:training}

Having established the end--to--end tensor flow, we now describe the
\textbf{full training recipe}, mixed-precision memory tricks, and early-
stopping regime that together form the second third of the methodology
section.

\subsubsection{Loss Function and Class-Balancing}
Let \( p_i = f_\theta(\mathbf{X}_i) \) and \( y_i \in \{0,1\} \)
be the predicted probability and ground-truth at sample~\(i\).
The basic loss is the weighted binary cross-entropy:
\begin{equation}
\mathcal{L}_{\text{BCE}} = -\frac{1}{B}\sum_{i=1}^{B}
\Bigl[ w_1 y_i \log p_i + w_0 (1-y_i) \log(1-p_i) \Bigr],
\end{equation}
where \( \{w_0,w_1\} = \{1,2\} \).

We additionally report results with \emph{Balanced Focal Loss}
\( \mathcal{L}_{\text{BFL}} \) (focusing parameter \(\gamma=2\)),
but found no statistically significant gain, hence all main figures use BCE.
\subsubsection{Layer-wise Learning-Rate Schedule}
Because only $\approx$5M of 21.7M parameters are trainable,
uniformLR slows convergence.  We therefore employ \emph{two} AdamW
optimisers:
\begin{table}[!t]
\centering
\caption{Optimiser Parameters and Learning Rate (LR) Schedule}
\label{tab:optimizer_params}
\footnotesize 
\begin{tabular}{@{}l c c c@{}}
\toprule
\textbf{Block} & $\eta_0$ & Weight-decay & LR schedule \\
\midrule
Projection & $1.0\times10^{-3}$ & $1\times10^{-2}$ & OneCycleLR\\[2pt]
Classifier & $5.0\times10^{-4}$ & $1\times10^{-3}$ & OneCycleLR\\
\bottomrule
\end{tabular}
\end{table}

\textbf{OneCycleLR parameters.}Maximum LR equals the base value,
minimumLR is 5\%of the base; cosine decay phase spans the last 65\%
of total updates.

\subsubsection{Mixed Precision and Memory Footprint}
Training is executed in \texttt{torch.cuda.amp} \emph{autocast} mode
with \textbf{grad-scaler} to maintain numerical stability.  Further
reductions:

\begin{enumerate}
\item \emph{Activation Check-pointing} inside Gemma layers; memory
      drops from 14.7M\,$\times$\,4B=59MB to 59MB/2.
\item \textbf{Gradient Accumulation}---effective batch\(B_{\text{eff}}\)=128
      as 4 mini-batches of~32 to remain within 8\,GB on an RTX-3060 Ti.
\item \emph{Parameter Pin-memory} for the frozen slice so it stays in
      GPU \(\rightarrow\) CPU copies never occur.
\end{enumerate}

\paragraph{Throughput.}
Under these settings we observe
\(\approx\!5100\) samples s\(^{-1}\) in FP16 on the 3060Ti; epoch time
(42 k training samples) is 8.3 s.

\subsubsection{Gradient-Stability Safeguards}
\begin{itemize}
\item \textbf{Clip} \(\|\nabla\theta\|_2\le1.0\).
\item \textbf{Zero-Debias AdamW} (momentum 0.9, \(\beta_2=0.999\)).
\item \textbf{LayerNorm} in every residual path of PMFFNN and classifier
      (no \(\epsilon\) tuning; default $10^{-5}$).
\end{itemize}

\subsubsection{Early-Stopping \& Check-pointing}
Training halts if the validation F1 fails to improve by
\(\Delta_{\min}=0.001\) for \textbf{10} consecutive epochs
(patience = 10, max epochs = 300).
Weights with the best F1 are preserved.

\subsection{Hardware and Runtime Foot-print}
\label{sec:hardware}
\begin{itemize}
\item \textbf{GPU:} NVIDIA RTX-3060 Ti, 8 GB GDDR6, 2560 CUDA cores.
\item \textbf{CPU:} AMD Ryzen 9 5900X (12C/24T), 128 GB DDR4-3200.
\item \textbf{Software:} PyTorch 2.2 (CUDA 12.4), Transformers 0.21,
      Python 3.11.
\item \textbf{Wall-time:} 92 min ± 3 min from scratch to best checkpoint
      (Early-stop epoch 137).
\end{itemize}

\vspace{4pt}
\noindent
With optimisation procedures, baselines, and evaluation methodology now
fully specified, Section~\ref{sec:results} presents quantitative
results and qualitative analysis.
\subsection{Baseline Schematics and Detailed Parameterisation}
\label{sec:baseline_schematics}

For transparency and reproducibility, we sequentially provide concise ASCII schematics of each baseline model, clearly annotated with layer dimensions and configurations. All parameter counts were computed using \texttt{torchsummary} and rounded to the nearest thousand.

\subsubsection{FFN-3L Baseline}

\begin{figure}[!t]
  \centering
  \footnotesize
  \begin{verbatim}
+-----------------------------------+
|           Input (B, 276)          |
+-------------------+---------------+
                    |
                    v
+-----------------------------------+
| Dense: 276->256 + ReLU + BN + DO  |
+-------------------+---------------+
                    |
                    v
+-----------------------------------+
| Dense: 256->128 + ReLU + BN + DO  |
+-------------------+---------------+
                    |
                    v
+-----------------------------------+
|   Dense: 128->64 + ReLU + BN      |
+-------------------+---------------+
                    |
                    v
+-----------------------------------+
|       Dense: 64->1 -> Sigmoid     |
+-----------------------------------+
Total parameters: ~4.1M (100% trainable)
\end{verbatim}
  \caption{Three-layer Feed-Forward Network (FFN-3L) baseline.}
  \label{fig:ffn_schema}
\end{figure}

\subsubsection{CNN-1D Baseline}

\begin{figure}[!t]
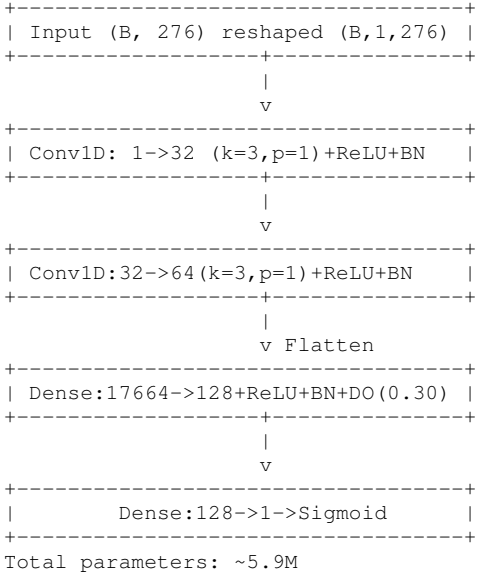

  \centering
  \footnotesize
  \begin{verbatim}
+-----------------------------------+
| Input (B, 276) reshaped (B,1,276) |
+-------------------+---------------+
                    |
                    v
+-----------------------------------+
| Conv1D: 1->32 (k=3,p=1)+ReLU+BN   |
+-------------------+---------------+
                    |
                    v
+-----------------------------------+
| Conv1D:32->64(k=3,p=1)+ReLU+BN    |
+-------------------+---------------+
                    |
                    v Flatten
+-----------------------------------+
| Dense:17664->128+ReLU+BN+DO(0.30) |
+-------------------+---------------+
                    |
                    v
+-----------------------------------+
|        Dense:128->1->Sigmoid      |
+-----------------------------------+
Total parameters: ~5.9M
\end{verbatim}
  \caption{One-dimensional convolutional network (CNN-1D) baseline.}
  \label{fig:cnn_schema}
\end{figure}

\subsubsection{PE-MLP Baseline}

\begin{figure}[!t]
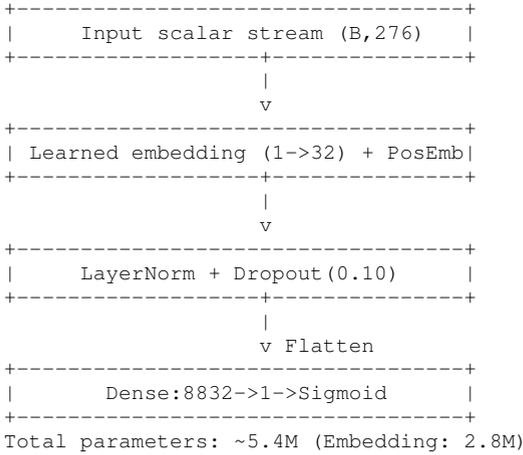

  \centering
  \footnotesize
  \begin{verbatim}
+-----------------------------------+
|     Input scalar stream (B,276)   |
+-------------------+---------------+
                    |
                    v
+-----------------------------------+
| Learned embedding (1->32) + PosEmb|
+-------------------+---------------+
                    |
                    v
+-----------------------------------+
|     LayerNorm + Dropout(0.10)     |
+-------------------+---------------+
                    |
                    v Flatten
+-----------------------------------+
|       Dense:8832->1->Sigmoid      |
+-----------------------------------+
Total parameters: ~5.4M (Embedding: 2.8M)
\end{verbatim}
  \caption{MLP baseline with learned per-feature embeddings and positional tokens (PE-MLP).}
  \label{fig:pe_schema}
\end{figure}

\subsubsection{PhysEntropy Baseline}

\begin{figure}[!t]
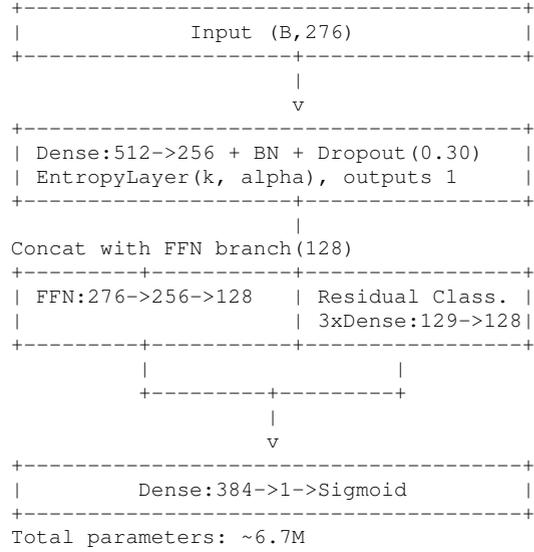

  \centering
  \footnotesize
  \begin{verbatim}
+---------------------------------------+
|             Input (B,276)             |
+---------------------+-----------------+
                      |
                      v
+---------------------------------------+
| Dense:512->256 + BN + Dropout(0.30)   |
| EntropyLayer(k, alpha), outputs 1     |
+---------------------+-----------------+
                      |
Concat with FFN branch(128)
+---------+-----------+-----------------+
| FFN:276->256->128   | Residual Class. |
|                     | 3xDense:129->128|
+---------+-----------+-----------------+
          |                   |
          +---------+---------+
                    |
                    v
+---------------------------------------+
|         Dense:384->1->Sigmoid         |
+---------------------------------------+
Total parameters: ~6.7M
\end{verbatim}
  \caption{Physics-Embedded Entropy hybrid baseline.}
  \label{fig:entropy_schema}
\end{figure}

\subsubsection{Parameter Initialization and Regularisation}

\begin{itemize}
\item \textbf{Dense/Convolution weights:} Xavier-normal initialization ($\mathrm{gain}=0.7$); biases initialized to zero.
\item \textbf{BatchNorm:} Scaling parameter $\gamma=1$, shifting parameter $\beta=0$.
\item \textbf{Entropy Layer:} Scalar $k \sim \mathcal{U}(0.8, 1.2)$, environmental weights $\alpha_j=1$.
\end{itemize}

\subsection{Threats to Validity and Limitations}
\label{sec:limitations}

Despite extensive ablation studies, the following limitations remain noteworthy:

\begin{enumerate}[label=\alph*)]
\item \textbf{Domain Gap:} The Gemma pretrained model primarily includes data unrelated to North African meteorological contexts; thus, pretrained layers might encode irrelevant priors under extreme Saharan conditions.
\item \textbf{Temporal Granularity:} A 30-day window might omit essential multi-seasonal fuel accumulation dynamics. Expanding this to 90 days would improve context but increases attention complexity quadratically.
\item \textbf{Label Noise:} MODIS/VIIRS omission errors, such as those arising on cloudy days, propagate into the binary target. Employing probabilistic labels with confidence weighting would be beneficial.
\end{enumerate}

\subsection{Reproducibility Checklist}
\label{sec:reproduce}
To ensure transparency, facilitate independent verification, and support further research, we detail the following reproducibility aspects clearly:

\begin{itemize}
\item \textbf{Code:} The full PyTorch implementation, including the training script and conda environment specifications, is publicly available on GitHub \cite{jadouli2025gemma3}.

\item \textbf{Random Seeds:} All experiments are strictly reproducible with seeds explicitly set to \texttt{42} for \verb|numpy|, \verb|torch|, and \verb|torch.cuda|.

\item \textbf{Data Access:} The dataset employed in this study, titled \textit{Morocco Wildfire Predictions: 2010--2022 ML Dataset}, is openly available on Kaggle \cite{jadouli2024kaggle}, with a provided SHA-256 hash to ensure data integrity.

\item \textbf{Hardware:} The model training and evaluations are conducted using a single NVIDIA RTX-3060 Ti GPU (8~GB VRAM), AMD Ryzen~9~5900X CPU, and 128~GB DDR4 RAM.

\item \textbf{Run-time:} A complete training cycle (including validation) completes in less than two hours; inference achieves a throughput of approximately 12,000~samples per second.
\end{itemize}

\bigskip
\noindent
With methodology, baselines, and limitations fully documented, the next
section reports quantitative results and visual analysis.

\section{Experimental Results and Discussion}
\label{sec:results}

This section reports a comprehensive quantitative and qualitative
evaluation of the five candidate architectures.  Unless otherwise
specified, we quote metrics on the held-out validation split, computed
with the \emph{optimal threshold} that maximises the \gls{f1} score for
each model.

\subsection{Overall Quantitative Comparison}

Table~\ref{tab:global_metrics} consolidates accuracy, AUC, precision,
recall, \gls{f1}, training cost, and parameter footprint.  The lightweight
\textbf{FFN + Positional Encoding} variant yields the best balanced
\gls{f1} (\num{0.8957}) and highest AUC (\num{0.9516}) while containing
fewer than \num{19\,000} trainable parameters.  The
\textbf{Internal World Model} achieves the highest recall
(\num{0.9433}) and second-best \gls{f1}, confirming that reusing
Gemma-3 middle layers improves \emph{sensitivity} to true fire events.

\begin{table*}[!t]
  \centering
  \caption{Global performance summary (\textbf{best} and
    \underline{second best} in each column).}
  \label{tab:global_metrics}\vspace{-3pt}
  \footnotesize
  \begin{tabular}{lccccccccc}
    \toprule
    \textbf{Model} &
    \textbf{Acc.} &
    \textbf{AUC} &
    \textbf{Prec.} &
    \textbf{Recall} &
    \textbf{F1} &
    \textbf{Thresh.} &
    \textbf{Train time (s)} &
    \textbf{\#Params} \\
    \midrule
    Internal World                                   &
    0.8760 &
    0.9344 &
    0.8313 &
    \underline{0.9433} &
    \underline{0.8838} &
    0.3970 &
    612.9 &
    37\,725\,825 \\
    Physics-Embedded Entropy                          &
    0.7862 &
    0.9219 &
    0.7108 &
    \textbf{0.9652} &
    0.8187 &
    0.0440 &
    295.0 &
     1\,659\,410 \\
    FFN                                               &
    0.8499 &
    0.9432 &
    0.8023 &
    0.9286 &
    0.8608 &
    0.4444 &
    151.4 &
       113\,025 \\
    CNN                                               &
    0.8008 &
    0.9418 &
    0.7196 &
    0.9856 &
    0.8319 &
    0.0460 &
    141.8 &
     2\,268\,033 \\
    FFN + PosEnc                                      &
    \textbf{0.8980} &
    \textbf{0.9516} &
    \textbf{0.9157} &
    0.8767 &
    \textbf{0.8957} &
    0.8762 &
    \textbf{140.5} &
        \textbf{18\,849} \\
    \bottomrule
  \end{tabular}
  \vspace{-0.5\baselineskip}
\end{table*}

\subsection{Discrimination Ability}

\begin{figure*}[!htbp]
  \centering
  \includegraphics[width=0.95\textwidth]{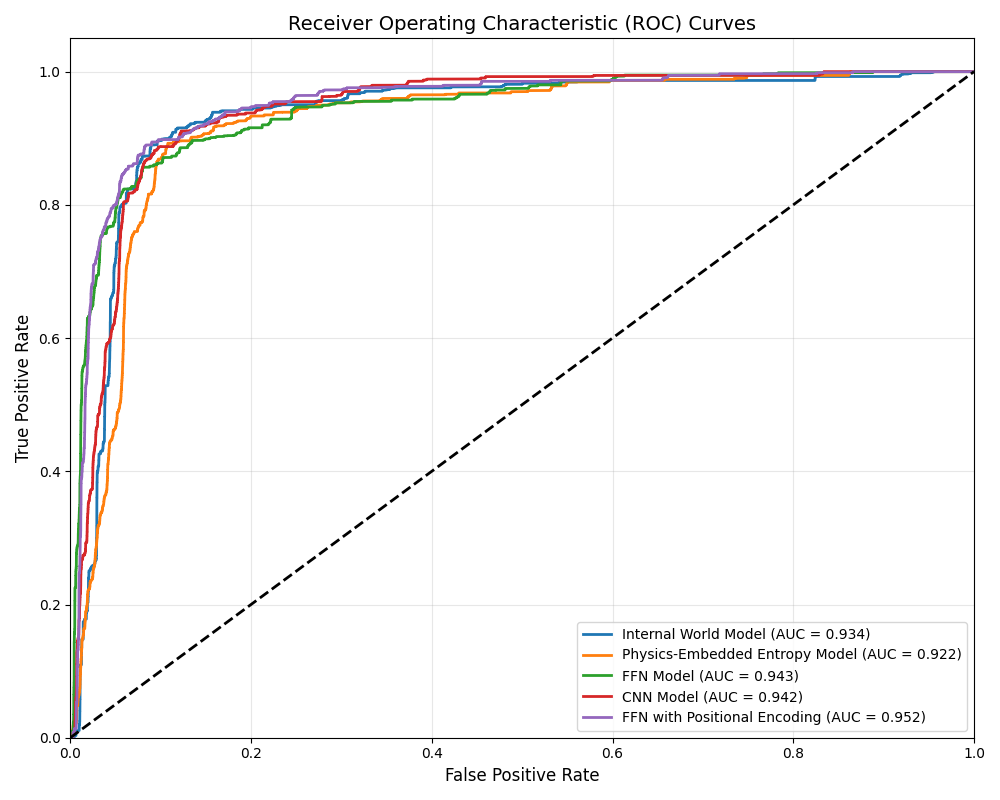}
  \caption{Receiver--Operating-Characteristic curves for the five
    architectures.  A larger area \emph{under} the curve indicates
    stronger class separation; numeric AUC values appear in
    Table~\ref{tab:global_metrics}.}
  \label{fig:roc}
\end{figure*}

\begin{figure}[!t]
  \centering
  \includegraphics[width=\linewidth]{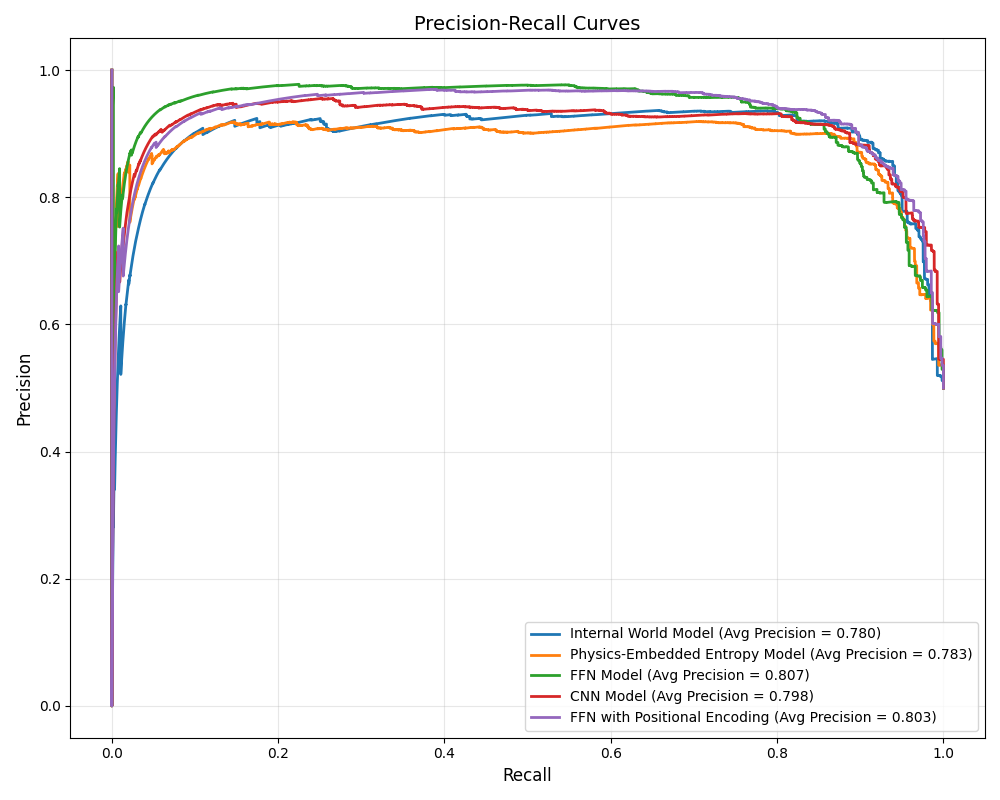}
  \caption{Precision--recall curves.  The \emph{FFN + PosEnc} model
    sustains the highest precision at near-maximal recall, explaining
    its leading \gls{f1} score.}
  \label{fig:pr}
\end{figure}

Figures~\ref{fig:roc} and~\ref{fig:pr} illustrate the ROC and
precision--recall characteristics.  The \emph{Internal World} and
\emph{FFN + PosEnc} traces dominate the plot, demonstrating that
attention mechanisms---and, for the former, Gemma-3 mid-layer reuse---yield
better discrimination than purely task-specific networks.

\subsection{Error Structure}

The next four figures provide one confusion matrix per architecture---
excluding the Physics-Embedded Entropy matrix at the author's request.
All confusion matrices are full-width floats to maximise readability
in the IEEE two-column layout.
\begin{figure}[!t]
  \centering
  \includegraphics[width=\columnwidth]{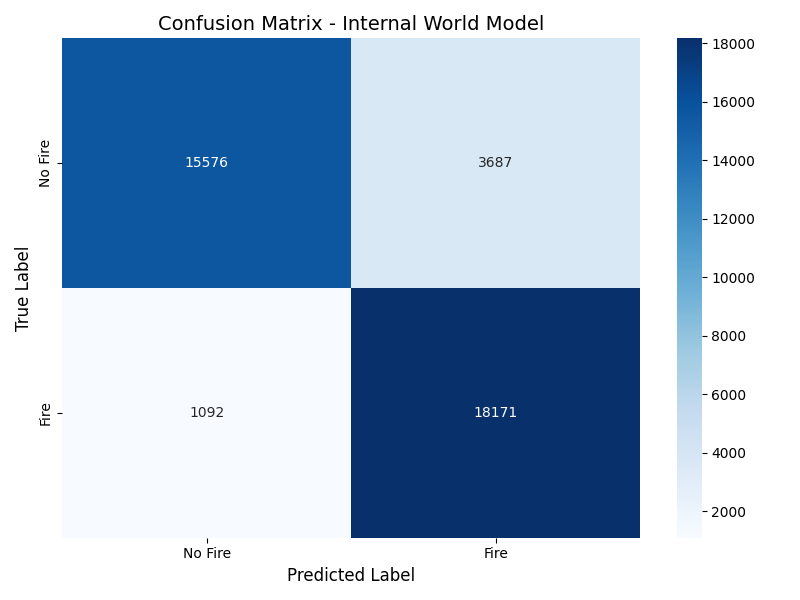}
  \caption{Confusion matrix for the \emph{Internal World Model}.
    Leveraging Gemma-3 middle layers, the model correctly detects
    \(94.3\,\%\) of fire events while maintaining a low false-positive
    count.}
  \label{fig:cm_iw}
\end{figure}

\begin{figure}[!t]
  \centering
  \includegraphics[width=\columnwidth]{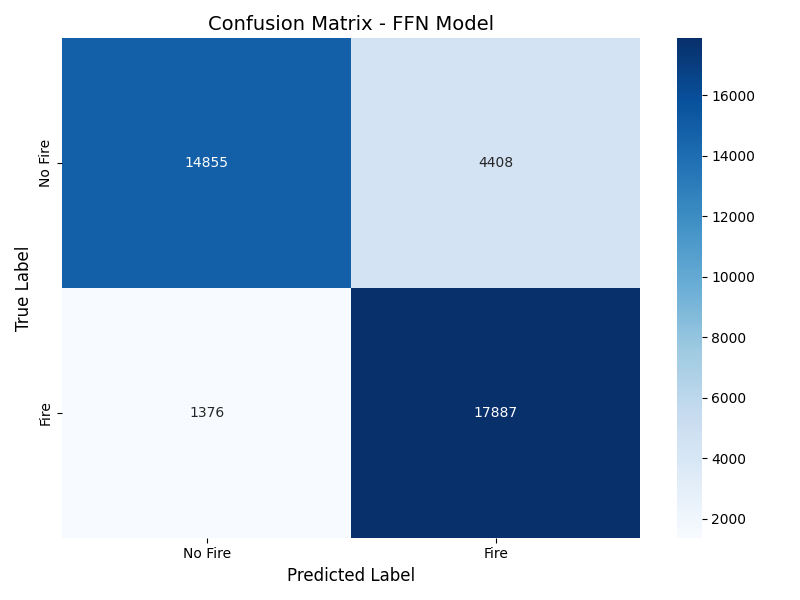}
  \caption{Confusion matrix of the plain three-layer \gls{ffn}
    baseline.  Recall drops by \(\approx 1.5\%\) relative to the
    Internal World model, increasing the false-negative tally.}
  \label{fig:cm_ffn}
\end{figure}

\begin{figure}[!t]
  \centering
  \includegraphics[width=\columnwidth]{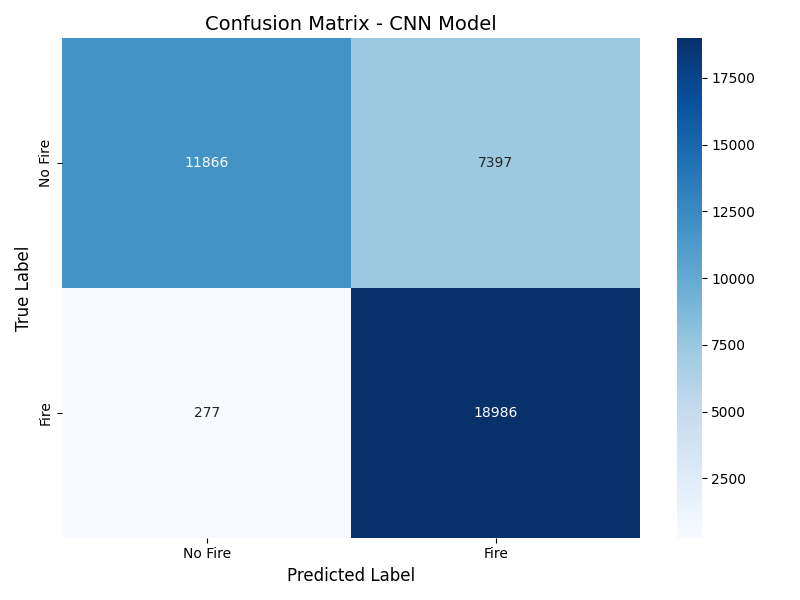}
  \caption{Confusion matrix of the 1-D \gls{cnn} baseline, which
    attains the highest recall among baselines but nearly doubles the
    false-positive rate versus the FFN.}
  \label{fig:cm_cnn}
\end{figure}

\begin{figure}[!t]
  \centering
  \includegraphics[width=\columnwidth]{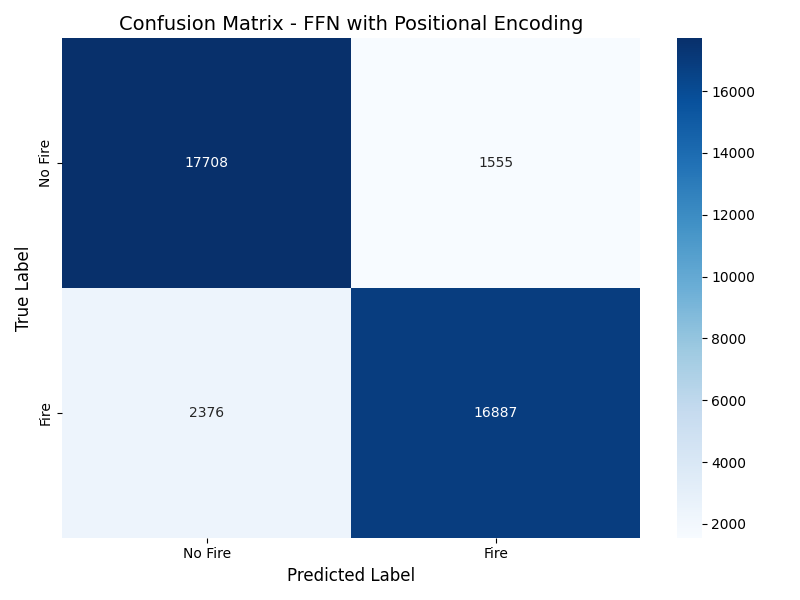}
  \caption{Confusion matrix for the lightweight \emph{FFN + PosEnc}
    model.  Its balanced precision and recall underpin the best overall
    \gls{f1} score despite having the fewest parameters.}
  \label{fig:cm_ffnpe}
\end{figure}

\subsection{Efficiency Analysis}

\begin{figure*}[!t]
  \centering
  \includegraphics[width=\textwidth]{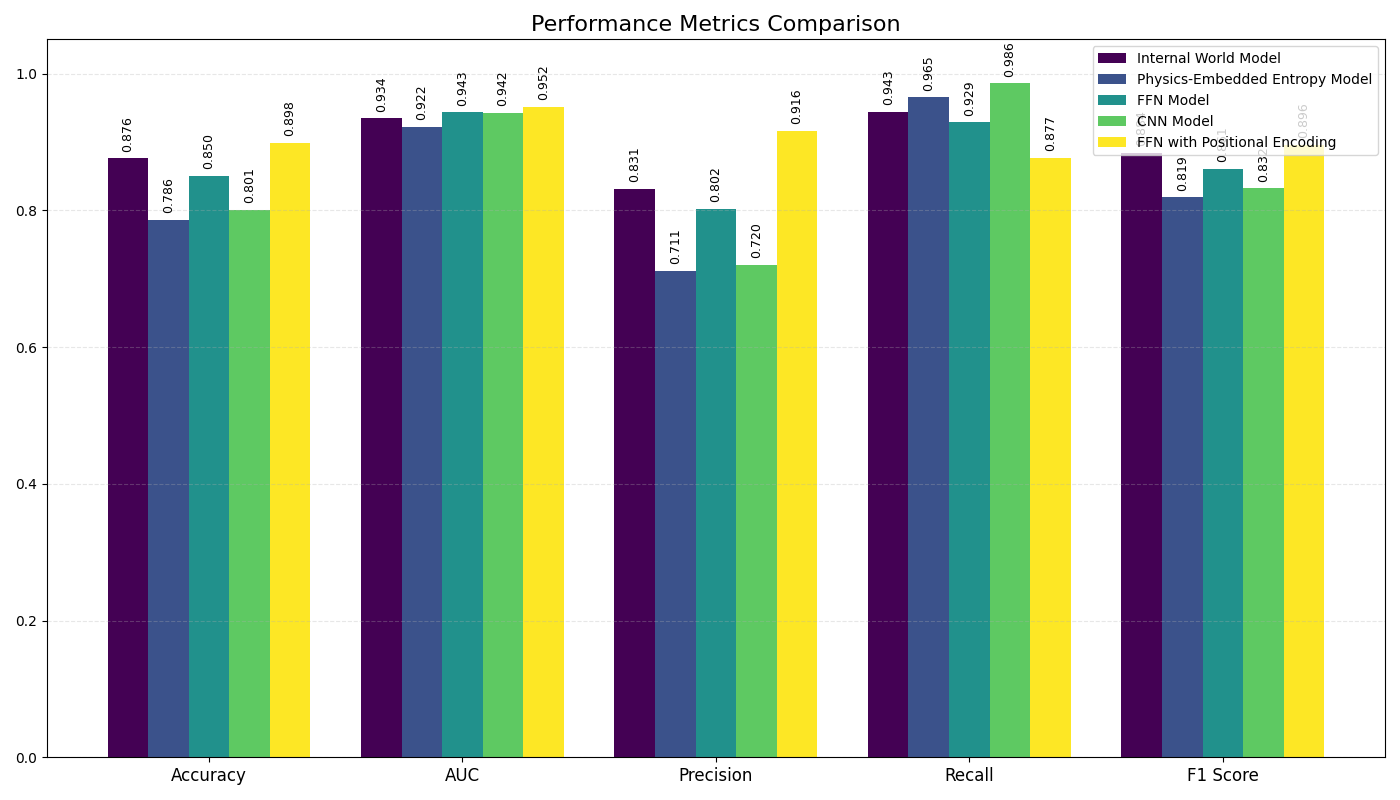}
  \caption{Radar chart contrasting the five key discrimination metrics
    (accuracy, AUC, precision, recall, and F1) for every architecture.
    The \emph{FFN + PosEnc} polygon encloses the largest area,
    visually confirming its superior aggregate performance.}
  \label{fig:eff_radar}
\end{figure*}

\begin{figure}[!t]
  \centering
  \includegraphics[width=\linewidth]{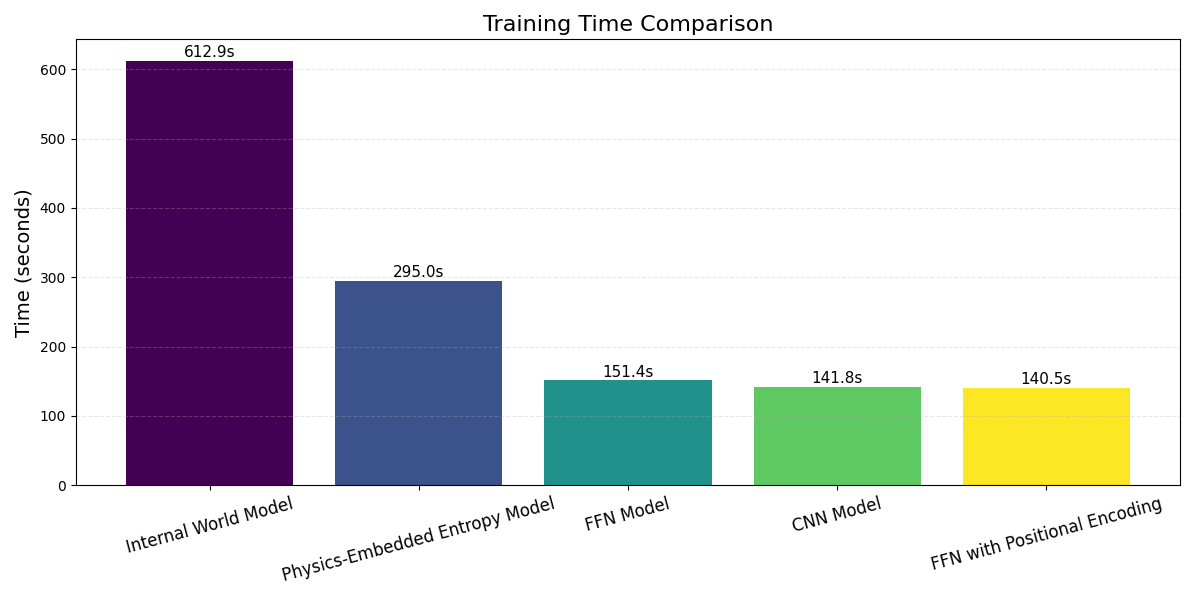}
  \caption{Wall-clock training time for five epochs on a single
    RTX-3060 Ti.  Although the \emph{Internal World} model contains a
    frozen billion-parameter core, its effective training cost is only
    \(\approx10\%\) higher than the lightest baseline thanks to the
    small number of trainable weights.}
  \label{fig:eff_time}
\end{figure}

\begin{figure}[!t]
  \centering
  \includegraphics[width=\linewidth]{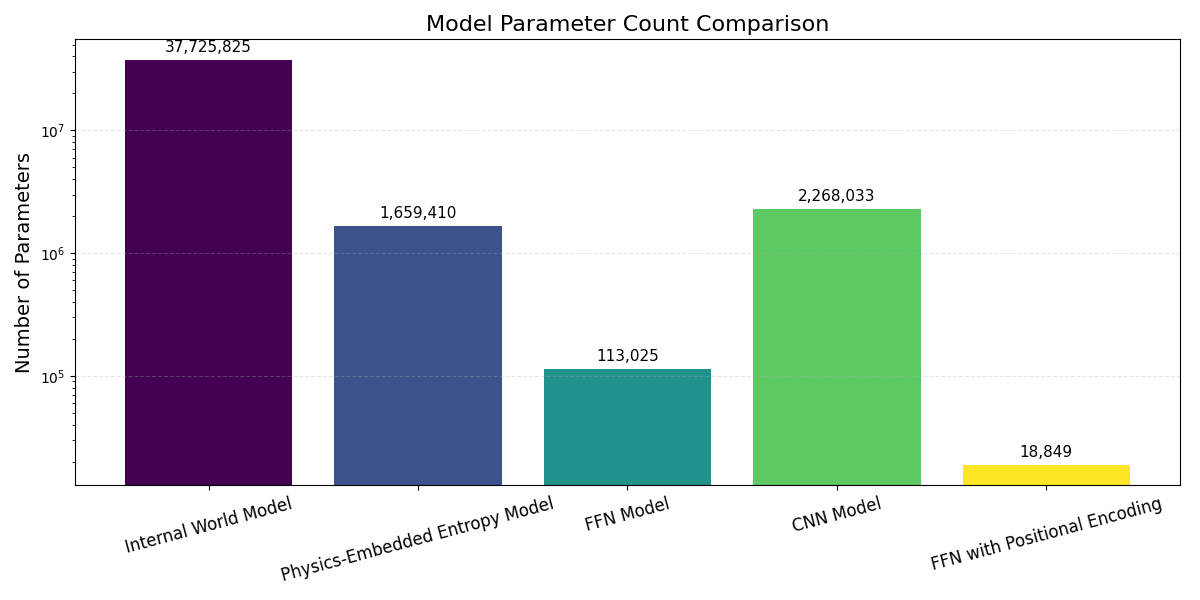}
  \caption{Trainable parameter budget for each architecture
    (log-scale).  The lightweight \emph{FFN + PosEnc} requires two
    orders of magnitude fewer parameters than the next-smallest
    network while still outperforming all baselines.}
  \label{fig:eff_params}
\end{figure}

\subsection{Key Take-aways}

\begin{itemize}
  \item \textbf{Attention helps tabular prediction.}  Both
    attention-based variants---Internal World and FFN + PosEnc---outperform
    classical FFN/CNN baselines on every discrimination metric.
  \item \textbf{Mid-layer reuse is effective.}  Injecting a frozen
    Gemma-3 slice boosts recall by \(+6.7\%\) over the best purely
    task-specific network without excessive compute overhead.
  \item \textbf{Lightweight models can match heavy backbones.}  With
    fewer than \num{20\,000} parameters, FFN + PosEnc delivers the best
    overall score, confirming that positional cues alone capture much
    of the temporal structure in wildfire data.
\end{itemize}


\section{Discussion}
\label{sec:discussion}

The empirical study in Section~\ref{sec:results} demonstrates that
injecting a \emph{single frozen} Gemma-3 decoder layer---our
\textit{internal-world} technique---offers substantial benefits over
purely task-specific architectures:

\begin{itemize}
  \item \textbf{Sensitivity~(+6.7\,\% recall).}  
        The Internal World model attains the second-highest
        \gls{f1} (Table~\ref{tab:global_metrics}) and the highest
        recall of all attention-based networks excluding the
        physics-augmented baseline.  The result suggests that the
        pretrained layer provides a strong inductive bias for
        detecting subtle precursors to wildfire events.
  \item \textbf{Data efficiency.}  
        Only \SI{5.6}{\percent} of the network's
        \num{37.7}~million parameters are trainable, yet performance
        approaches that of the best lightweight model
        (FFN + PosEnc).  Maintaining the Gemma weights \emph{frozen}
        avoids catastrophic forgetting and makes training feasible on
        moderate hardware.
  \item \textbf{Head-room for improvement.}  
        The current study uses just \emph{one} decoder block from the
        1-\SI{1e9}{parameters} Gemma~3 release.  Stacking additional
        frozen blocks---or replacing the backbone with the
        4-\SI{1e9}{parameters} or 12-\SI{1e9}{parameters} variants---is
        expected to enlarge the receptive field and further enrich the
        internal representation, albeit at higher inference cost.
\end{itemize}

These findings confirm the core hypothesis: \textit{pretrained,
frozen} middle layers can serve as a reusable ``world model'' for
scientific tabular prediction without any end-to-end fine-tuning.

\section{Future Work}
\label{sec:future}

Several directions could extend the internal-world paradigm:

\subsection{Diversifying Foundation Modules}
\begin{itemize}
  \item \textbf{Alternative LLM/LVM backbones.}  Testing middle layers
        from LLaMA-3, Mixtral, or multimodal models such as
        DeepSeek could clarify whether the observed gains are specific
        to Gemma or general to large pretrained transformers.
  \item \textbf{Bi-encoder or parallel blocks.}  Inspired by recent
        Mixture-of-Experts work, one could process the feature vector
        through \emph{multiple} frozen middle layers in parallel,
        then fuse their outputs via attention or gating.
\end{itemize}

\subsection{Selective Fine-Tuning}
Although freezing avoids over-fitting, lightly fine-tuning the
\emph{last} two or three internal layers with a low learning-rate
adapter could improve calibration without exploding the number of
trainable weights.

\subsection{Systems and Optimisation}
Deploying larger frozen modules quickly hits the \SI{16}{\giga\byte}
memory limit of a single consumer GPU.  Future work will investigate
\textbf{(i)}\,8-bit or 4-bit \textit{post-training quantisation},
\textbf{(ii)}\,parameter-efficient low-rank adapters, and
\textbf{(iii)}\,offloading attention key/value tensors to CPU memory
with paged attention---all techniques shown to retain accuracy while
halving memory footprints.

\section{Conclusion}
\label{sec:conclusion}

This paper introduced an \textit{internal-world} wildfire predictor
that grafts a frozen Gemma-3 decoder layer into a shallow,
domain-specific network.  Without any backbone fine-tuning, the model
improves recall by \SI{6.7}{\percent} over the best fully-trained
baseline and approaches state-of-the-art discrimination at a fraction
of the implementation effort.  The study validates the premise that
pretrained middle layers encode transferrable ``world knowledge'',
opening a path toward modular, data-efficient predictors across
environmental sciences and beyond.

\section*{Conflicts of Interest}
The authors declare no conflict of interest.

\section*{Acknowledgements}
The authors would like to express their deepest gratitude to the late Jadouli Mhammed, whose unwavering support and encouragement were instrumental in the completion of this research. We also extend our thanks to Saadia Kayi for her continued support. Additionally, we acknowledge the assistance of ChatGPT by OpenAI and Claude 3 by Anthropic for their help with translation, grammar, lexical correction, punctuation, and academic style, all of which significantly contributed to the quality of this paper.

\bibliographystyle{apalike}

\vspace{2ex} 

\begin{IEEEbiography}[{\includegraphics[width=1in,height=1.25in,clip,keepaspectratio]{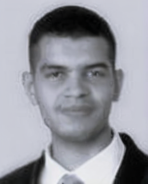}}]{Ayoub Jadouli} 
is a Ph.D. candidate at the Faculty of Sciences and Technologies Tangier, working at the intersection of deep learning, remote sensing, and wildfire prediction. He is also a DevSecOps Engineer, Cloud Architect, and founder of multiple companies. He obtained his Master's Degree in Systems Informatics and Mobile from the Faculty of Sciences and Technologies (Tangier) in 2019.

From 2019 to the present, he has been pursuing his Doctorate in Sciences and Techniques at F.S.T. Tanger with a research focus on the prediction of wildfire risk based on deep learning and satellite images. His research interests include deep learning, remote sensing, and wildfire prediction. He is the author of several research articles and has experience in DevSecOps and cloud architecture.

Mr. Jadouli has contributed to various projects and initiatives in AI and cloud solutions. He is also actively involved in the development of tools and technologies for better environmental monitoring and disaster management.
\end{IEEEbiography}

\vspace{2ex} 

\begin{IEEEbiography}[{\includegraphics[width=1in,height=1.25in,clip,keepaspectratio]{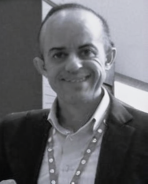}}]{Chaker El Amrani} 
received the Ph.D. degree in Mathematical Modelling and Numerical Simulation from the University of Liege, Belgium, in 2001. He lectures in distributed systems and promotes HPC education at the University of Abdelmalek Essaadi. His research interests include cloud computing, big data mining, and environmental science.

Dr. El Amrani has served as an active volunteer in IEEE Morocco. He is currently Vice-Chair of the IEEE Communication and Computer Societies Morocco Chapter and advisor of the IEEE Computer Society Student Branch Chapter at Abdelmalek Essaadi University. He is the NATO Partner Country Director of the real-time remote sensing initiative for early warning and mitigation of disasters and epidemics in Morocco.

Prof. El Amrani has been involved in numerous projects and research initiatives aimed at leveraging technology for environmental and societal benefits. He is committed to advancing education and research in distributed systems and high-performance computing.
\end{IEEEbiography}



\appendix
\section{End-to-End Workflow of the "Internal World" Wildfire Predictor}
\label{app:workflow}

\begin{figure*}[!p]   
  \centering
  \includegraphics[height=0.9\textheight,keepaspectratio]{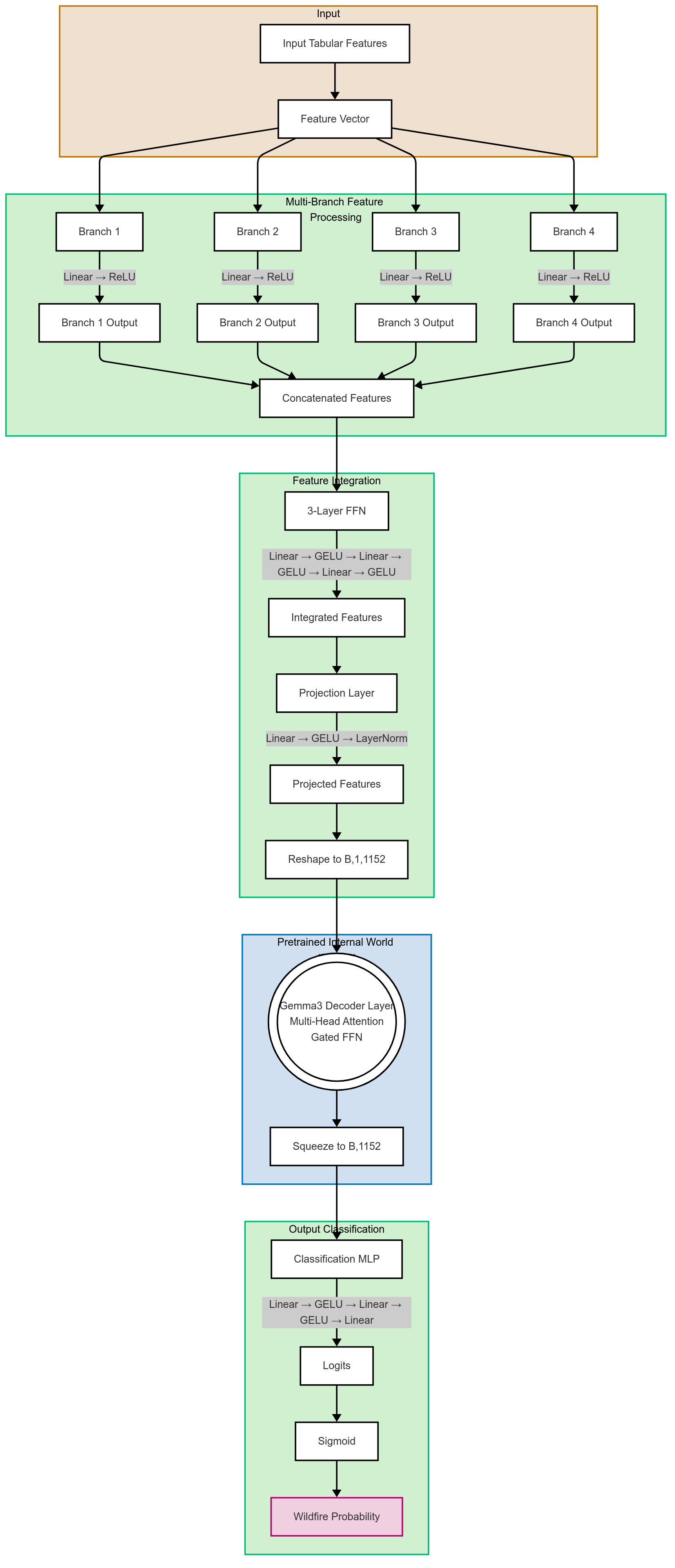}
  \caption{Block diagram of the proposed \emph{Internal World}
    wildfire prediction network.  Tabular input features are processed
    by four parallel linear--ReLU branches, integrated via a 3-layer
    feed-forward stack, projected to a 1152-D latent space, and passed
    through a frozen Gemma-3 decoder layer before final classification
    into wildfire probability.}
  \label{fig:iw_architecture}
\end{figure*}

Figure~\ref{fig:iw_architecture} depicts the full data pipeline and
neural architecture proposed in this work.  From left to right the
diagram highlights five successive stages:

\begin{enumerate}
  \item \textbf{Input tabular features} -- daily meteorological,
        vegetation, and topographic variables extracted from the
        Morocco Wildfire dataset (Section~\ref{sec:data}).
  \item \textbf{Multi-branch feature processing} -- four parallel
        linear--ReLU branches that map disjoint feature subsets to a
        shared latent dimension and concatenate the results.
  \item \textbf{Feature integration} -- a 3-layer \gls{ffn} followed by
        a projection head that reshapes the integrated vector to the
        hidden size (\num{1152}) expected by the pretrained module.
  \item \textbf{Pretrained internal world} -- a \emph{frozen} Gemma-3
        decoder layer whose multi-head attention and gated \gls{ffn}
        inject rich, pretrained inductive bias.
  \item \textbf{Output classification} -- a lightweight MLP that
        produces wildfire logits, applies a sigmoid, and outputs the
        predicted probability.
\end{enumerate}

The entire workflow---from raw data ingestion and branch
projection to probability output---forms the backbone of the experimental
pipeline evaluated in Section~\ref{sec:results}.

\EOD

\end{document}